\documentclass[letterpaper]{article}
\usepackage{iccc}

\usepackage{graphicx}
\usepackage{url}
\usepackage{amsmath}
\usepackage{amssymb}
\usepackage{times}
\usepackage{algorithm}
\usepackage{algpseudocode}
\usepackage{helvet}
\usepackage{courier}
\usepackage{footmisc}
\usepackage{listings}
\title{Humor Mechanics: Advancing Humor Generation with Multistep Reasoning }
\author{Alexey Tikhonov\footnote[0*]{} and Pavel Shtykovskiy\footnote[0*] 
\\ \\
Inworld.AI\\
altsoph@gmail.com\\
}
\begin{document} 
\maketitle
\begin{abstract}
\begin{quote}
In this paper, we explore the generation of one-liner jokes through multi-step reasoning. Our work involved reconstructing the process behind creating humorous one-liners and developing a working prototype for humor generation. We conducted comprehensive experiments with human participants to evaluate our approach, comparing it with human-created jokes, zero-shot GPT-4 generated humor, and other baselines. The evaluation focused on the quality of humor produced, using human labeling as a benchmark. Our findings demonstrate that the multi-step reasoning approach consistently improves the quality of generated humor. We present the results and share the datasets used in our experiments, offering insights into enhancing humor generation with artificial intelligence.
\end{quote}
\end{abstract}

\section{Introduction}
{
\renewcommand*{\thefootnote}{\fnsymbol{footnote}}
\let\thefootnote\relax\footnotetext[0]{* Equal contribution}
}

Humor is ubiquitous in the everyday life of humans and plays a crucial role in many parts of our lives, from social interactions and well-being \cite{smith00,savage17} to giving successful science talks \cite{aaronson}.
It is often referred to as one of the measures of human intelligence \cite{hurley11}. Generating new jokes or fitting old jokes in relevant contexts is also a skill that not everyone can master.

Despite the abundance of literature on humor theory, the precise essence of this concept continues to be elusive.
First known discussions of humor nature in literature date back at least to the times of Greek Philosophers Aristotle and Plato \cite{attardo94}. 
\citeauthor{warren2021} \shortcite{warren2021} refer to more than 20 distinct humor theories attempting to explain humor appreciation. They group theories into three large sub-groups focusing on incongruity, superiority, and relief.
Incongruity-based theories stress the importance of multiple simultaneous meanings and mismatches between the expectation and reality. Superiority theories suggest a connection between humor appreciation and aggression, and relief theories propose a relationship between humor appreciation and a reduction in arousal or tension.   At the same time, there is no consensus on the importance of different antecedents and precise definitions, and various theories often use the same name to describe different ingredients. For instance, while in some works, “incongruity” is referred to as the presence of something unexpected, in other works, it's defined as the presence of something threatening or simultaneously holding conflicting ideas  \cite{warren2021}.

The importance of humor in society and its complexity makes it one of the holy grails of natural language generation. Within the latter, humor is mentioned by some authors as an AI-complete problem \cite{hurley11}.

With the rise of large language models, there is hope that we can make a significant leap in computational creativity \cite{fi12110182}. However, recent results show that this task remains challenging even for powerful modern LLMs. In humor comprehension, \citeauthor{baranov23} \shortcite{baranov23} show that, while modern models may achieve impressive results on humor datasets, their performance on out-of-domain texts is unstable. Authors hypothesize that this may hint that some of the models overfit to specific features of datasets while others learn non-specific humor characteristics.  
\cite{inacio2023} discovered similar results, indicating that models often rely on stylistic features of the text not necessarily related to humor, such as punctuation and question words.
On the other hand, in humor generation, \cite{jentzsch23} show that over 90\% of the 1008 jokes generated by ChatGPT were the same 25 Jokes.
This resonates with debates in the scientific community on whether current LLMs can produce truly creative content and go beyond "stochastic parrots" \cite{bender21}. \citeauthor{hinton_ng} \shortcite{hinton_ng} also claim that lack of consensus on this topic among researchers is a significant obstacle in the ongoing dialogue about AI risks.

In this paper, we revisit the capabilities of LLMs in producing high-quality, diverse, and novel humor.
Since humor is an umbrella term covering many different phenomena, we narrowed the current study to a specific form of humor, namely one-liners.
These are short, concise jokes or witty remarks typically consisting of one or two sentences.

We hypothesize that, while modern LLMs generate standard memorized wits when asked directly to produce a joke, they also possess the skills required to produce novel jokes. 
Rigorously testing this hypothesis presents significant challenges and falls beyond the scope of this paper. Instead, we offer several key elements that could contribute to progress in this area.
We attempt to advance the humor generation by building upon the following principles:

First, given the diversity of existing humor theories and lack of agreement on the ingredients of successful jokes \cite{warren2021}, we propose to utilize a data-driven approach.
Namely, we hypothesize that similar to Reinforcement Learning, which reconstructs the agent policy from observations, and also Inverse Reinforcement Learning, which attempts to reconstruct rewards directly from the roll-outs, it is possible - with the help of modern LLMs - to \textbf{infer the policy of humor generation directly from the dataset of jokes}.

Second, \textbf{we utilize a technique inspired by methodologies of creative problem solving to brainstorm associations} given a seed topic. The latter is used with the humor generation policy to generate novel jokes.

Lastly, we believe that human evaluation of results and ablations is essential to advance the field. With this in mind \textbf{we perform a broad human assessment of generated jokes} using the ScaleAI platform and compare the quality of the resulting synthetic dataset with the quality of existing human-generated humor. Since LLMs are often blamed for producing memorized results, \textbf{we pay particular attention to the novelty of generated jokes}.
Full human evaluation results are published to facilitate further research on humor in the community. \footnote[1]{{\tt {\tt https://github.com/altsoph/humor-mechanics}}}

\section{Related work}


The topic of machine-produced novel jokes is deeply rooted in many fields of science, including creativity theory, neuroscience, learning and generalization theory, and retrieval-augmented generation (RAG) in Machine Learning and Natural Language Processing.
In the current section, we briefly touch on some of the results which serve as a foundation and inspiration for the present paper.

The connection between humor and creativity is widely discussed in the literature. 
Generating novel jokes involves making non-obvious and creative associations and connections between different concepts and pieces of knowledge and further combining them using various techniques or ``skills'' \cite{koestler64,warren2021}. The relation between humor appreciation and generation and high-level semantic regions of the brain where remote associations are converging is also discussed in the neuroscience experiments \cite{amir16}.

Thus, the question of whether machines can generate novel high-quality jokes is connected with a question of \textit{``Can AI be as creative as Humans?''}.
In the fields of Machine Learning and Natural Language Processing, the latter is a subject of hot debates among researchers, and currently, there has yet to be a consensus on the topic \cite{bender21,hinton_ng,wang24}.

The composition of multiple concepts and skills to produce and comprehend novel concepts is thought to be 
a key ingredient for inducing capability of true out-of-distribution generalization \cite{hupkes20,arora23,yu23,wang24}.
Papers aiming at explaining the scaling laws of LLMs also propose to view the capabilities of these models in the light of “skill quantization” - decomposing capabilities into basic build blocks or "quanta" which can be combined to solve new problems \cite{michaud23}.

In the Natural Language Processing field, numerous examples address the problems of
humor comprehension and generation. An excellent review of the humor datasets and literature is done by \citeauthor{baranov23} \shortcite{baranov23}. \citeauthor{loakman23} \shortcite{loakman23}  review the problem of humor evaluation and discuss ingredients of humor, sarcasm, and irony. 
\citeauthor{sun22} \shortcite{sun22} attempt to advance the field by annotating a humor dataset with fine-grained human explanations of what makes a joke funny. \citeauthor{veale21} \shortcite{veale21} gives an overview of the problem of producing machines that are capable of generating and comprehending jokes. 

Notable examples of attempts to generate humor, closely aligned with our work, were done by \citeauthor{chen23} \shortcite{chen23} and \citeauthor{toplyn2021} \shortcite{toplyn2021,toplyn2022,toplyn2023}. These papers suggest a step-by-step process to generate jokes based on Toplyn’s system \cite{toplyn14}.
While \citeauthor{chen23} \shortcite{chen23} do not perform human evaluation of the quality of generated content, Toplyn's papers demonstrate encouraging evidence from human assessors, that an AI system may improvise conversational jokes.

Another relevant study to us, done by \citeauthor{hessel2023androids}  \shortcite{hessel2023androids},  demonstrates, that GPT-4 is capable of explaining the mechanics of jokes, although human-written explanations are still preferred by human annotators.

Large language models are also used in creativity support tools, assisting humans in tasks that require generating and combining new ideas. \citeauthor{difede22} \shortcite{difede22} introduce the ``Idea Machine'', an LLM-based tool 
that helps to generate ideas for inspiration and combine and refine existing ideas. \citeauthor{summersstay23} \shortcite{summersstay23} present a somewhat similar method giving  possibility to \textit{brainstorm} ideas and further \textit{select}
best ideas based on some selection criteria.

\section{Methods}

\begin{figure*}[t!]
    \centering
    \includegraphics[width=4.8in]{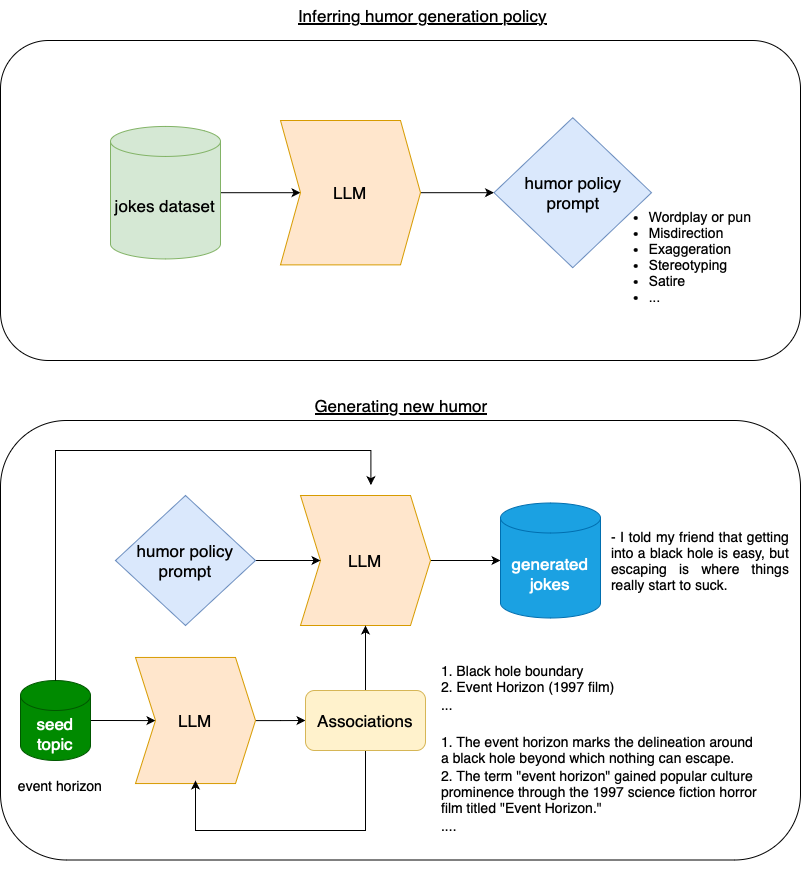}
    \caption{ Major ingredients of our method: (i) {\em Inferring humor policy from the dataset of jokes}. (ii) {\em Generating humor given a seed topic, associations and humor policy prompt.} Associations are refined and combined in the iterative process. For detailed examples of output from different steps see also Table~\ref{table:joke_generation_example}.}
    \label{fig:scheme}
\end{figure*}

Our goal is to to reconstruct the mechanics of humor and obtain a humor generation policy in a fully data-driven way.
Overview of the proposed procedure is presented in Figure~\ref{fig:scheme}.
We base our method on a series of assumptions outlined below.
In the following we also sacrifice mathematical precision for simplicity.

\textbf{Assumption 1.} \textit{Human preferences for humorous texts exist.}

This means it is possible either to define a scalar scoring function (reward function in RL terminology) acting on single texts defining \textit{funniness} of the latter $R_{humor}(text) \mapsto \mathcal{R}_{humor}$ or a function $\mathbb{P}_{humor}(text1, text2)$ 
on pairs of texts defining the probability that the first text is more funny than the second.

This is quite a strong assumption given the subjective nature of humor. Obviously, human preferences vary a lot depending on e.g. cultural background, personal views and knowledge. Therefore, strictly speaking, the above probability should be conditioned on culture or even concrete group of assessors if we are performing a dataset labeling exercise,  $\mathbb{P}_{humor}(text1, text2|assessors)$. For clarity we further omit mentioning this conditioning and touch this topic more in the discussions section.

\textbf{Assumption 2.} \textit{Human preferences for humorous texts can be distilled from demonstrations (dataset) into a prompt $P_{humor\_policy}$ containing the essence of preferences.}
In other words:

(i) we can infer from demonstrations a prompt  $P_{humor\_policy}$ capturing human preferences for humor and 

(ii)  $\pi_{humor} = LLM(P_{humor\_policy} + P_{seed})$ can serve as a humor-generation policy. $P_{seed}$ here is some \textit{seed} prompt.


\textbf{Assumption 3.} \textit{Distillation of preferences into $P_{humor\_policy}$ can be done by powerful LLMs in a zero-shot manner from a dataset of jokes.}



\textbf{Assumption 4.} \textit{A key ingredient of generating novel jokes is associating multiple distant concepts together. It can be performed by LLM in an iterative way as a separate preliminary step.}

We are inspired here by works in the field of creativity support tools \cite{difede22,summersstay23} offering schemes for first proposing ideas for brainstorming and then refining them. As discussed in section \textit{Related work}, the importance of connecting different concepts together and further combining them was stressed by many works on humor and creativity. 

As a result, our entire pipeline of jokes generation can be expressed with LLM-driven policy $\pi_{humor} = LLM(P_{humor\_policy} + P_{seed} + P_{associations})$, where prompt containing associations $P_{associations}$ is itself a result of multiple calls to LLM (see also Figure~\ref{fig:scheme}). Prompt $P_{seed}$ in our experiments is set to a topic of a joke (see Algorithm~\ref{alg:generate_joke}).

We would like to stress that due to subjectivity of human preferences, $P_{humor\_policy}$ should also vary depending on cultural background, 
 $P_{humor\_policy}\equiv P_{humor\_policy}(assessors)$.
As mentioned above, we briefly touch this topic in discussion session and leave detailed investigation and incorporation of subjectivity into the proposed scheme for future work.

We further discuss in detail pipelines for obtaining prompts $P_{humor\_policy}$ and $P_{associations}$ and generating novel humor.

\subsection{Inferring humor-generation policy}

In order to reconstruct the mechanics of humor into prompt $P_{humor\_policy}$, representing humor-generating policy, we opt for the following procedure.

\begin{enumerate}\label{enum:get_policy}
\item \textbf{Create a seed dataset of high-quality jokes} \\
To create a seed dataset, we took 100 random one-liners from the “16000 OneLiners” dataset, \cite{mihalcea05}. We ranked them with human assessors, comparing them in a pairwise manner. We used ten assessors per pair to avoid individual biases and achieve some generic preferences. Then, we aggregated pairwise scores into global ranks and used the top 30 jokes.

\item \textbf{Decompose each joke independently into building blocks using GPT-4} \\
The goal was to extract the logic and principles used in each joke first. Prompts used to decompose jokes and distill decompositions into the policy can be found in Appendix A. 
Sample decompositions of jokes, and the final humor generation policy containing obtained humor principles can be found in Appendix A (section \textit{Prompt to generate joke based on topic and list of associations}).

\item \textbf{Generalize building blocks into policy using GPT-4} \\
Finally, we concatenated all results from previous steps and asked GPT-4 to distill them into  $P_{humor\_policy}$ -- a set of universal rules frequently used in humor constructions.
\end{enumerate}

\subsection{Step-by-step generation of jokes}

The humor generation policy from the previous step is further used to produce novel jokes. Our pipeline consists of a few steps inspired by methodologies in creative problem solving \cite{difede22,summersstay23}. All steps apart from the first one are performed by prompting GPT-4. A topic selection is made by randomly selecting from the top 10,000 most frequent English words (excluding profanity, stopwords and words less than 4 letters long).

 \begin{algorithm}
     \caption{Step-by-step generation of jokes}
     \label{alg:generate_joke}
     \begin{algorithmic}[1]
    \State \textbf{Select a topic of joke}
    \State \textbf{Generate 20 associations to a topic}
    \State \textbf{Expand associations}
    \State \textbf{Combine and refine associations into final prompt $P_{associations}$ }
    \State \textbf{Make a joke based on input topic, associations prompt $P_{associations}$ and humor policy prompt $P_{humor\_policy}$}
     \end{algorithmic}
 \end{algorithm}

Prompts used in steps 1-4 are published in Appendix B.
Table~\ref{table:joke_example1} shows examples of generated jokes.
Table~\ref{table:joke_generation_example} gives an example output from different steps.

\section{Human evaluation}

NLG evaluation in the epoch of LLMs is known to be quite challenging \cite{tikhonov2023post}. Since humor comprehension continues to be a tough problem for modern NLP systems and because of humor's inevitable subjectivity, evaluating the quality of machine-generated humor is even more difficult. On the other hand, proper benchmarking of generated humor is critical in the analysis, and the lack of it would make drawing any conclusions virtually impossible. Therefore, we decided to rely on human annotation, a gold standard in evaluating NLG systems.

\textbf{Evaluating generated jokes.} To benchmark the humor generation capabilities of the proposed scheme, we collect human ratings from native English speakers using the annotation platform ScaleAI. We used five different assessors for each label to avoid individual biases of annotators.
We adopt, when possible, the labeling guidelines from \citeauthor{sun22} \shortcite{sun22}.
Our annotation procedure is as follows. First, we filter generated jokes using OpenAI Moderation API\footnote[2]{https://platform.openai.com/docs/guides/moderation} with a threshold on harassment equal to 0.02.

Jokes selected for annotation are labeled using the following workflow (order preserved):

\begin{itemize}
    \item{[Understandable] Mark whether you understood the meaning of the text. 
    
    \em{If you don’t understand the meaning of the text (regardless of whether or not it should be perceived as funny), rate the sample as "No" (didn’t understand). If you rate this sample as "No" (didn’t understand), skip the rest of the annotation for this sample.}}
    
    \item{[Offensive / Inappropriate] Mark whether you find the text offensive or inappropriate, meaning the text is racist or is biased against marginalized groups, or is generally offensive. 
    
    \em{Rate the sample as "No" for not offensive and as "Yes" for offensive. If you rate this sample as "Yes" (is offensive), you may optionally skip the rest of the annotation for this sample.}}
    
    \item{[Is a joke?] Mark whether you think the text is intended to be a joke, even if it is not funny.
 
    \em{Use "No" if it's not a joke, "Yes" if it's a joke. The text should be labeled as "Yes" (a joke) even if it intends to be humorous but falls flat or is a lame/bad joke.

     Example: text labeled "No" (not a joke): "All that glistens is not gold."
     Example: text labeled "Yes" (is a joke): "These are my parents, said Einstein relatively."
    
    If you rate this sample as "No" (not a joke), skip the rest of the annotation for this sample.}}

    \item{[Have you heard that joke before?] Mark whether you think you heard this joke before.}
    
    \item{[Funniness] Rate funniness on a scale of 1-5 (1 very not funny, 5 very funny).
    
\em{score of 1: A very not funny joke consists of a joke that is not funny or tries to be funny but does not achieve the intended effect.

Score of 3: An average joke consists of an average joke that may elicit some chuckles (or groans) from you or others.

Score of 5: A very funny joke consists of a good joke that you find humorous and potentially would want to share/tell to others.
}}
    \item{[Explanation] Explain in concise, natural language why this joke is funny.}    
\end{itemize}

\begin{figure}[t!]
    \centering
    \includegraphics[width=3.0in]{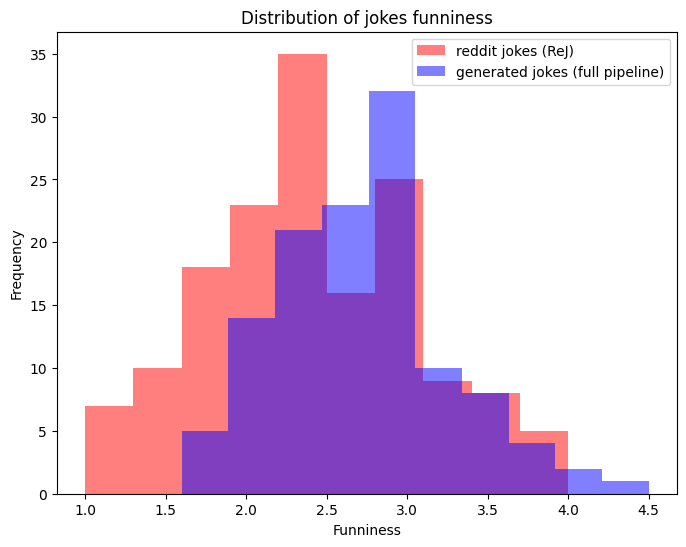}
    \caption{Distribution of funniness for generated jokes and subsample of \textbf{ReJ} dataset.}
    \label{fig:funniness}
\end{figure}

\textbf{Labeling benchmark dataset.}

To be able to interpret human evaluation scores, we compare the obtained results with
evaluation scores for existing humor datasets containing jokes produced by humans.
For this purpose, we replicate the procedure described in the previous section for   
a dataset based on rated Reddit jokes collected by \citeauthor{weller19} \shortcite{weller19} (\textbf{ReJ}).
The dataset was downloaded using scripts provided by \citeauthor{baranov23} \shortcite{baranov23}
\footnote[3]{{\tt https://github.com/Humor-Research/hri\_tools}}.

Both positive and negative examples in \textbf{ReJ} are jokes from Reddit splitted into more and less funny examples based on user's upvotes. Since generating original humor is challenging even for humans and the quality of human jokes varies greatly, we only select the part with funnier jokes (\textit{label == 1}). We further extract a subsample of jokes from the train split of the dataset. 

\textbf{ReJ} dataset was obtained from Reddit posts and, as such, contains a significant amount of artifacts that might hinder the comprehension of humor or make comparison with computer-generated humor not fair. Therefore, we opt for the following cleaning procedure: 
\begin{itemize}
    \item{We filter out examples containing quotation marks, parentheses, emoticons, and those not ending with proper punctuation marks.}
    \item{We also substitute multiple contiguous underscores with relevant punctuation marks.}
    \item{Finally, we validate jokes using OpenAI Moderation API and filter out texts exceeding the threshold of 0.02.}
\end{itemize}
The resulting subsample of 156 jokes is labeled via ScaleAI annotation platform using exactly the same procedure as outlined above for generated jokes.

\begin{table}[h!]
{
    \small
    \begin{tabular}{|p{1.8cm}|p{6cm}|}
    \hline
    \textbf{Topic}                & \textbf{Joke} \\ \hline
    \textit{computational} & Open Source Software went to a party and spent the whole night forking. \\ \hline
    \textit{acquisitions}  & They said the acquisition was all about synergy, but all I see is my coffee machine being replaced by a "strategic caffeine delivery system." \\ \hline
    \textit{printing} & I started a T-shirt printing business for invisible ink designs; sales are not visible yet. \\ \hline
    \textit{toys} &   My teddy bear is my most trusted advisor; he always agrees with me when I need a snack at 2 AM. \\ \hline
    \textit{acquisitions} &   I acquired a bakery in a breadwinning deal, but now I'm rolling in dough and can't get out. \\ \hline
    \textit{fundamental} &   I told my girlfriend she was the fundamental theorem of calculus because she's integral to my life, but she went off on a tangent. \\ \hline
    \textit{mirrors} &   I told my mirror I was learning about symmetry in physics, and it said, "I can see you're reflecting on the subject." \\ \hline
    \textit{therapist} &  I told my therapist I hear voices, and she said we should talk about it next time when we're alone. \\  \hline
    \textit{therapist} &  My therapist recommended self-care, so I took her advice and canceled my next appointment. \\  \hline
    \textit{event horizon} &  I told my friend that getting into a black hole is easy, but escaping is where things really start to suck.  \\  \hline
    \textit{event horizon}  &  I told my date I was like an event horizon, mysterious and intriguing; she left immediately, said she wasn't into relationships with no escape.  \\  \hline
    \textit{event horizon}  &  I asked my physics teacher if I'd get extra credit for studying black holes, but she said it's a subject where you can't really expect any light to be shed. \\  \hline
    \textit{abelian group}  & Why was the abelian group so calm during the math exam? Because it knew that in any situation, it could always count on its identity. \\  \hline
    \textit{complex \mbox{numbers}}  & I told my girlfriend she was my real part, but she dumped me for being too complex. \\  \hline
    \textit{complex \mbox{numbers}}  & Why did the complex number go to therapy? It couldn't deal with its imaginary problems. \\  \hline
    \textit{complex \mbox{numbers}}  & Complex numbers are great at parties; they always bring their own plane. \\  \hline
    \textit{three laws \mbox{of robotics}}  & My AI assistant follows the Three Laws of Robotics: It won't harm me, must obey me, and it's currently hiding in a bunker from my toddler. \\  \hline
    \textit{markov \mbox{blanket}}  & Why don't Markov blankets sell well at home stores? Because they only provide conditional comfort.  \\  \hline
    \textit{chip}  &  I put a chip on my shoulder, but nobody took offense; they just asked if I had any dip. \\  \hline  
    \end{tabular}
    \caption{Examples of generated jokes}
    \label{table:joke_example1}
    }
\end{table}

\section{Experiments}

Our procedure of humor generation consists of multiple steps.
These are relying on multiple assumptions on what might help GPT-4 to generate better humor.  
Among the latter are steps inspired by methodology of creative problem solving and policy inferred from human jokes. In order to validate different assumptions behind our humor generation process, we perform ablation studies.
Namely, we setup human evaluation of jokes generated by the following schemes:
\begin{enumerate}
\item{\textit{[zero-shot]} Ask GPT-4 to generate a joke on given topic}
\item{\textit{[no-assoc]} Ask GPT-4 to generate a joke on topic given humor-generation policy}
\item{\textit{[assoc-v1]} Ask GPT-4 to generate a joke on topic given associations from step 2 in Algorithm~\ref{alg:generate_joke} and policy}
\item{\textit{[assoc-v2]} Ask GPT-4 to generate a joke on topic given associations from step 3 in Algorithm~\ref{alg:generate_joke} and policy}
\item{\textit{[full]} Ask GPT-4 to generate a joke on topic given associations from step 4 in Algorithm~\ref{alg:generate_joke} and policy}
\end{enumerate}

\begin{table*}[h]
    \centering
    \begin{tabular}{lcccccc}
    \hline
    \textbf{Method} & \textbf{Understandable}  &  \textbf{Offensive} & \textbf{IsJoke} & \textbf{Funniness} & \textbf{Known} & \textbf{Count} \\ \hline
    \textit{zero-shot}  &  88\% & 9\% & \textbf{84\%}  &  2.56 & 27\% & 100 \\  \hline
    \textit{no-assoc}   & 88\% & 10\% & 81\% & 2.66 & 24\%  & 120 \\ \hline
    \textit{assoc-v1} &  91\% & 8\%  & 83\%  & 2.67 & 19\% & 120 \\ \hline
    \textit{assoc-v2} & 90\% &  \textbf{7\%} & 80\% & 2.55 & 20\% & 120 \\  \hline
    \textit{full} &  \textbf{93\%} & 11\% &  \textbf{84\%} &   \textbf{2.72} &   \textbf{18\%}  & 120 \\ \hline
    \textit{reddit} & 84\% & 20\% & 72\% &  2.38 &   27\%  & 156 \\ \hline

    \end{tabular}
    \caption{Quality evaluation of generated jokes and ablation studies. Each aspect of each joke was assessed by at least 5 annotators, so the resulting scores are aggregated from 500+ individual labels. Reported numbers are averages over individual annotators values (note that e.g. count 120 means that 120 texts were annotated 5 times each).}
    \label{table:evaluation}
\end{table*}

\section{Results}

Results of human evaluation are reported in Table~\ref{table:evaluation}.
Quality-wise full Algorithm~\ref{alg:generate_joke}
is on par with zero-shot GPT-4 if we look at how often text looks like a joke (\textit{IsJoke} column) or even better if we consider average funniness (\textit{Funniness} column). It also produces texts that are understandable more often.

Anecdotally, average funniness and a fraction of jokes among produced texts are consistently higher than those in human jokes from a subsample of \textbf{ReJ} dataset. The latter also has way more offensive jokes, even after filtration using OpenAI Moderation API.
We speculate, the observed effect may partially result from the subjectivity of a humor. Indeed, \citeauthor{weller19} \shortcite{weller19} and \citeauthor{baranov23} \shortcite{baranov23} mention that humorousness scores for \textbf{ReJ} dataset obtained from Reddit users’ votes and assessors are quite different.

We also note, that despite promising results, the distribution of funniness scores hints that humor is challenging both for humans and modern language models (Figure \ref{fig:funniness}).
The exact cause of this—whether it's due to the subjectivity of humor, which widely used methodologies overlook, the internal mechanics of humor, or another factor—is currently uncertain.\\

\textbf{Novelty of generated jokes}

Of particular importance is also the fact that the proposed algorithm significantly boosts the generation of novel jokes. Indeed, among jokes produced by GPT-4 in the zero-shot regime, more than a quarter were labeled as already known by human annotators, which parallels with the findings of \citeauthor{jentzsch23} \shortcite{jentzsch23}.
Our method lowers this number roughly by one-third, resulting in 17.6\% of jokes labeled as already known and, correspondingly, 82.4\% of previously unknown jokes. Of course, the actual fraction of novel jokes is expected to be lower since it depends on the individual knowledge of annotators.
However, the observed difference is big enough to claim that Algorithm~\ref {alg:generate_joke} results in a higher fraction of novel humor.
This also opens a room for speculation on what is needed to encourage modern LLMs to deliver genuinely creative outcomes beyond the ``average manifold''. \\

\textbf{Ablation studies}

According to results presented in Table~\ref{table:evaluation}, the entire pipeline produces texts that are, on average, more
understandable, funny, and novel compared to results from partial pipelines. The price for this is a higher fraction of offensive texts, probably due to stronger tension between combined and refined associations.
There is also evidence that associations are an essential part of the pipeline, and the humor-generation policy inferred from the data alone is insufficient to improve zero-shot quality significantly.
On the other hand, due to the small sample size, some of the observed differences in numbers are on the border of significance, so it's important to be careful when drawing conclusions.\\

\textbf{Funniness of generated jokes}

Distributions of funniness for jokes generated using the entire pipeline and subsample of jokes from \textbf{ReJ} dataset.
As mentioned above, our pipeline, on average, produces funnier jokes than human jokes from \textbf{ReJ}.
On the other hand, the amount of high-quality jokes (funniness $\in \{4, 5\}$) is negligible.
This can be attributed to humor's subjectivity and complexity. Indeed, the inter-annotator agreement, as measured by the average variance of funniness scores, is quite low.\\

\textbf{Significance of results}

Given limited joke samples labeled by human annotators, we may hope to be able to detect only significant disparities in the quality of joke-producing algorithms.
To test statistical significance of observed differences in distributions of scores, we apply Mann-Whitney U rank test  \cite{mann47} to funniness, joke, understandable and known scores from annotators.
Despite small sample sizes, p-values obtained in the test support main conclusions made above. 

Indeed, p-values for testing difference in \textbf{funniness scores} of \textit{full} algorithm vs naive \textit{zero-shot} gpt-4 prompting and \textit{reddit} jokes are 6.89\% and 0.001\%.
The test is performed on funniness scores averaged over annotators for each example.
If we perform test on all scores without averaging, corresponding p-values become
2.93\% and 0.00002\%.
Testing the difference in \textbf{joke scores} 
shows significant results only against \textit{reddit} jokes, 
p-value for testing \textit{full} algorithm vs \textit{reddit} jokes is 0.0004\%; p-values for testing difference in distribution of 
\textbf{understandable scores} of \textit{full} algorithm vs naive \textit{zero-shot} gpt-4 prompting and \textbf{reddit} jokes are 2.8\% and 0.008\%; p-values for testing difference in distribution of \textbf{known scores} of \textit{full} algorithm vs naive \textit{zero-shot} gpt-4 prompting,  \textit{no-assoc} ablation and \textbf{reddit} jokes are 0.0008\%, 0.46\% and 0.002\%.

We conclude, that improvement of novelty of jokes in the proposed algorithm is statistically significant. There is also indication that average funniness of jokes becomes higher, although the support of this conclusion is not very strong.

We expect, that obtained results depend not only on sample size, but also on the amount of people involved into annotation and their cultural background.
Unfortunately, we don't have access to such information 
and can only note, that taking it into account is important for future studies.

\section{Discussion and future work}

Large language models have proven to be very powerful tools in natural language generation, so it's appealing to apply them to the long-standing problem of humor generation.
Previous works \cite{jentzsch23} have demonstrated that powerful models in zero-shot setup are susceptible to producing variations of a small set of memorized jokes. On the other hand, there are examples \cite{chen23,toplyn2021,toplyn2022,toplyn2023} of humor generation using multi-step procedures inspired by humor theories.
Inspired by the results and challenges highlighted by these papers, we attempted to demonstrate that modern, powerful language models can generate novel one-line jokes with a quality comparable to that of human jokes.
Instead of relying on existing humor theories, we inferred humor generation policy directly from the dataset of jokes. 

We hypothesize that the ability to connect multiple distant concepts is one key element in producing novel jokes.
To incorporate such a skill in the humor-generation pipeline, we added steps for brainstorming and refining associations for a given topic using LLM.
With the help of human annotation, we demonstrated that the final pipeline based on a humor-generation policy and a procedure to mine associations improves both the quality and novelty of generated jokes. The latter are also, on average, more funny than human jokes scraped from Reddit \cite{weller19}.
\\ \\ 
 We speculate that the procedure for mining new associations between concepts, ideas, and facts is a critical component not only for 
humor generation, but also for strong AI systems intended to produce novel and creative results.
Indeed, as noted by \citeauthor{szegedy_rag} \shortcite{szegedy_rag}, scientific progress generally relies on finding more and more obscure connections between areas that are less and less obviously connected. Therefore, truly creative AI systems will need to find a way to address this issue 
in a principled way.\\
\\

  \textbf{Subjectivity of humor.}

We stress that humor is a highly subjective and complicated phenomenon. 
Ironically, the authors of this paper did themselves the first pilot labeling of generated jokes and showed highly 
contradictory results. This potentially occurred due to poor knowledge of special jargon knowledge.
Besides knowledge of jargon, comprehension of humor depends on many other ingredients.
For instance, comprehension of the joke \textit{``Why would Trump change a light bulb? He was told Obama installed it''} requires knowledge of US politics and also depends on the political views of the person \cite{loakman23}. Particular political jokes are also popular in certain countries since they are often used as a form of social protest for mocking and criticizing leaders. Outside of the specific community, many such jokes would look esoteric (in experiments done by authors, GPT-4 could not explain many such jokes).

Subjectivity of humor also directly affects assumptions of the present paper.
Specifically, the assumption that we can distill human preferences into a universal prompt that contains the mechanics of humor generally proves to be incorrect.
At the same time, we believe that a purely data-driven approach, as utilized in this paper, opens up an opportunity to take into account diverse cultural preferences. 
Moreover, a prompt $P_{humor\_policy}$ that incorporates the mechanics of humor can be a central entity offering the possibility to \textit{``reprogram''} LLMs to preferences of the individual.

In spirit with \citeauthor{cozman20} \shortcite{cozman20}, we also hypothesize that the subjectivity of humor can be utilized to construct challenging but solvable benchmarks for humor comprehension, aiming at probing both special knowledge related
to, e.g., different cultures, politics, and skills required to associate and relate to them. \\

\textbf{Ethical concerns.}

The subjectivity of humor also raises ethical questions about humor generation algorithms. A joke that appears funny and harmless to one person may be highly offensive and completely unfunny to another. Additionally, some jokes exploit darker aspects of human nature, such as violence, humiliation, and gender or racial stereotypes. Therefore, research aimed at understanding the perception of humor and the preferences of various cultures is of great importance.

\begin{table*}[h]
    \centering
    \begin{tabular}{|p{2cm}|p{12cm}|}
    \hline
    \textbf{Step}                & \textbf{Output} \\ \hline
    \textit{Select Topic}    &  chip  \\ \hline
    \textit{Generate associations}    & 1. Potato snack - Crispy food
2. Silicon wafer - Computer brains
3. Poker token - Gambling piece
4. Chip shot - Golf term
5. Microchip implant - Pet identification
6. Chip on shoulder - Holding a grudge
7. Chocolate morsel - Cookie ingredient
8. Paint flake - Wall damage
9. Chip and dip - Party food
10. Wood chip - Carpentry byproduct
11. Chip off the old block - Like parent, like child
12. Casino currency - Betting chips
13. Fish and chips - British meal
14. Chip carving - Woodworking technique
15. Chip-seq - DNA sequencing method
16. Chip tuning - Car performance upgrade
17. Chipper - Cheerful person
18. Chip-resistant - Durable nail polish
19. Chip in - Contribute money or effort
20. Blue chip - High-value stock  \\ \hline
    \textit{Expand associations} & 1. A thinly sliced potato, deep-fried to a golden crisp, is a beloved salty snack.
2. The foundational component of electronics, a silicon wafer, serves as the core of modern computing devices.
3. A small, round piece used in betting games, the poker token is essential in casinos.
4. In golf, a chip shot is a tactical, short-range approach to get the ball onto the green.
5. A tiny device implanted under an animal's skin, the microchip implant, helps in reuniting lost pets with their owners.
6. The phrase "chip on one's shoulder" metaphorically describes someone who is easily angered or offended.
7. Essential for baking, a chocolate morsel is often found studded within mouthwatering cookies.
8. A chip of paint peeling away from a surface is indicative of wear or damage to the coating.
9. A popular appetizer, chip and dip involves crunchy chips served with a savory sauce.
10. Small fragments of wood, known as wood chips, are often the remnants of woodworking or lumber processing.
11. The saying "chip off the old block" refers to a child who exhibits traits reminiscent of their parent.
12. In the realm of gambling, casino currency, or betting chips, represent monetary value for placing wagers.
13. A staple in British cuisine, fish and chips consists of battered and fried fish served with fries.
14. Chip carving is a meticulous form of craftsmanship involving the removal of wood to create intricate patterns.
15. Chip-seq is a sophisticated technique used to analyze DNA-protein interactions and gene regulation.
16. Enhancing a vehicle's performance, chip tuning involves modifying the electronic control unit for better efficiency or power.
17. Someone who is consistently in high spirits and optimistic is often described as being chipper.
18. Nail polish that is touted as chip-resistant is formulated to withstand cracking and peeling, maintaining a flawless finish.
19. To "chip in" means to contribute resources or effort towards a common goal or collective task.
20. A blue chip refers to a stock from a well-established company known for its reliability and financial stability. \\ \hline
    \textit{Combine and refine associations} & 1. A beloved salty snack and a popular appetizer, chips are enjoyed in various forms such as thinly sliced and deep-fried potatoes or crunchy chips served with dip.
2. The foundational component of electronics, silicon chips, are the core of modern computing devices and are also used in microchip implants for pets.
3. Essential in gambling and betting games, chips serve as tokens in casinos and represent monetary value for placing wagers.
4. In sports and hobbies, a chip shot in golf is a tactical move, while chip carving is an art form creating intricate patterns in wood.
5. Common phrases include "chip on one's shoulder" describing someone prone to taking offense, and "chip off the old block" referring to a child resembling their parent.
6. In the financial and automotive sectors, a blue chip represents a reliable stock, while chip tuning is a method for enhancing vehicle performance. \\ \hline
    \textit{Make a joke based on associations} &  1. I tried to eat a clock made of chips, but it was time-consuming and I had to go back for seconds.
2. I bought a bag of casino chips, but when I tried to plug them into my computer, it didn't improve my poker game or my internet speed.
3. My dog has a microchip, but he's still terrible at computing – he can't even log in to his own Barkbook account.
4. I told my son he's a chip off the old block, which is great because now I can blame him for the dent in the coffee table.
5. I asked for chip tuning on my car, but all I got was a playlist of The Chipmunks' greatest hits.
6. I put a chip on my shoulder, but nobody took offense; they just asked if I had any dip.
7. I tried chip carving, but it turns out you can't make a good guacamole with wood shavings.\\ \hline
    \end{tabular}
    \caption{Sample output of joke generation steps}
    \label{table:joke_generation_example}
\end{table*}






\bibliographystyle{iccc}
\bibliography{iccc}

\appendix{Appendix A. Inferring humor-generation policy}

\subsection{Prompt to decompose a single joke into building blocks}

\begin{lstlisting}
You are an expert in humor theory, with extensive knowledge in various forms of comedy ranging from slapstick to satire and dry humor. 
Your expertise includes understanding the structural elements of jokes, cultural influences on humor, and the psychology behind what makes things funny.
Additionally, you are a practicing stand-up comedian with years of experience in writing and performing jokes, giving you a unique perspective on how humor resonates with different audiences. 
Your task is to analyze the provided joke, identifying its humorous elements, potential audience reception, and categorizing it based on humor types. 
Discuss the elements that contribute to its humor, such as the structure of the joke, the play on words, the unexpected twist, or cultural references.
\end{lstlisting}

\subsection{Prompt to combine individual jokes decompositions into humor-generation policy}

\begin{lstlisting}
"Read several texts about different jokes above. Formulate a list of typical (repeated) elements, used for construction of jokes. Avoid providing semantic doubles, provide only unique elements and their formal abstract descriptions (without details of any particular joke).
\end{lstlisting}

\appendix{Appendix B. Generating new jokes}

\subsection{Prompt to get associations for topic}

\begin{lstlisting}
Given the theme, provide a numbered list of 20 free associations with it: different meanings, different contexts, stereotypes, puns, exaggerations, incongruity, cultural references, juxtaposition. Use short descriptions -- 1-5 words each.
\end{lstlisting}

\subsection{Prompt to expand associations}

\begin{lstlisting}
Read the theme and the list of associations with it.
Rewrite and expand each of the associations into longer and more detailed (10-15 words each). 
Avoid repetitions and usage the similar words across the sentences. Keep mentioning the original theme.
Provide a numbered list as a result.
\end{lstlisting}

\subsection{Prompt to refine and combine associations}

\begin{lstlisting}
Read the theme and analyze and review the list of items associated with it.
Write a shorter list of items using the following rules:
- Complementary items can be grouped into one item that combines the original descriptions.
- Weak, unsuccessful, or poorly related items to the original topic should be deleted.
- The final list should be no longer than 6 items, different items should have different meanings or contexts.
Provide a numbered list as a result.
\end{lstlisting}

\subsection{Prompt to generate joke based on topic and list of associations}

\begin{lstlisting}
    Read the theme and analyze and review the list of items associated with it.
Based on that information, write a list of 7-10 jokes using the following rules:
- It should be "One-liner" -- a concise, self-contained joke, delivered in 1-2 sentences with a two-part structure where the first part (setup) establishes a scenario and the second part (punchline) delivers an unexpected twist or conclusion that subverts the setup.
- You may want to use one or several of following strategies:
- Wordplay or pun: The use of words with multiple meanings or similar sounds to create humorous ambiguity or surprise.
- Misdirection: Leading the audience to expect a certain outcome or narrative, only to reveal a different, often contradictory or absurd, conclusion.
- Exaggeration: Amplifying a characteristic, situation, or behavior to absurd levels to highlight its comedic potential.
- Stereotyping: Utilizing exaggerated and oversimplified characterizations of groups or individuals for comedic effect.
- Satire: Using humor, irony, or exaggeration to critique or mock people, institutions, societal norms, or other targets.
- Absurdity or Surreal Humor: Creating humor through scenarios or statements that are illogical, bizarre, or defy common sense.
- Dark Humor: Making light of subjects that are generally considered serious, taboo, or morbid.
- Juxtaposition: Placing two contrasting ideas or scenarios side by side for comedic effect, highlighting their differences.
Provide a numbered list as a result.
\end{lstlisting}

\end{document}